\documentclass{article}
\usepackage{spconf,amsmath,graphicx}
\usepackage{multirow}
\usepackage{cite}
\usepackage{booktabs}
\usepackage{amssymb}
\usepackage{array}
\usepackage{threeparttable}

\usepackage{xcolor}
\usepackage{hyperref}
\newcommand{\proposed}{CoLLD}

\title{CoLLD: Contrastive Layer-to-layer Distillation\\for Compressing Multilingual Pre-trained Speech Encoders}
\name{
    Heng-Jui Chang$^{1,2,\ast}$, Ning Dong$^2$, Ruslan Mavlyutov$^2$, Sravya Popuri$^2$, Yu-An Chung$^2$
    \thanks{$^{\ast}$ Work done during an internship at Meta AI.}
}
\address{
    $^1$MIT CSAIL 
    $^2$Meta AI\\
    {
        \normalsize\texttt{hengjui@mit.edu, andyyuan@meta.com}
    }
}

\begin{document}
\ninept
\maketitle
\begin{abstract}
Large-scale self-supervised pre-trained speech encoders outperform conventional approaches in speech recognition and translation tasks.
Due to the high cost of developing these large models, building new encoders for new tasks and deploying them to on-device applications are infeasible.
Prior studies propose model compression methods to address this issue, but those works focus on smaller models and less realistic tasks.
Thus, we propose Contrastive Layer-to-layer Distillation~(CoLLD), a novel knowledge distillation method to compress pre-trained speech encoders by leveraging masked prediction and contrastive learning to train student models to copy the behavior of a large teacher model.
CoLLD outperforms prior methods and closes the gap between small and large models on multilingual speech-to-text translation and recognition benchmarks.
\end{abstract}

\section{Introduction}
\label{sec:intro}

Self-supervised learning~(SSL) for speech encoder pre-training benefits various speech processing tasks and outperforms conventional approaches~\cite{mohamed2022self}.
SSL methods leverage large unlabeled speech corpus to train deep neural networks to encode useful representations and succeed in applications like speech translation~\cite{seamlessm4t2023} and automatic speech recognition~(ASR)~\cite{zhang2023usm}.
However, powerful speech encoders usually have many parameters, making real-time or on-device speech processing less feasible.

Researchers propose model compression techniques to address the issues of large speech encoders.
The compressed SSL pre-trained encoders can be applied to various downstream tasks.
These approaches can be categorized into knowledge distillation~(KD) and parameter pruning.
In KD, a lightweight student model learns to predict hidden representations to mimic the large teacher model's behavior~\cite{chang2022distilhubert,lee2022fithubert,ashihara2022deep,wang2022lighthubert,huang2023ensemble,jang2023recycle,wang2023distilxlsr}.
DistilHuBERT~\cite{chang2022distilhubert} predicts multiple hidden layers in a HuBERT teacher~\cite{hsu2021hubert} using the student's output with separate prediction heads.
FitHuBERT~\cite{lee2022fithubert} and Ashihara et al.~\cite{ashihara2022deep} propose layer-to-layer~(L2L) KD that uses narrow and deep students to layer-wise distill the teacher's hidden representations.
In unstructured pruning, parameters with small values are set to zero~\cite{lai2021parp}, while structured pruning removes submodules from a model~\cite{peng2023structured,jiang2023accurate,wang2023task} to reduce the parameters but requires complicated implementation.
Other studies combine the above methods~\cite{peng2023dphubert} or techniques like layer-skipping~\cite{peng2023i3d} and low-bit quantization~\cite{yeh2022efficient}.

\begin{figure}[t]
    \centering
    \includegraphics[width=0.75\linewidth]{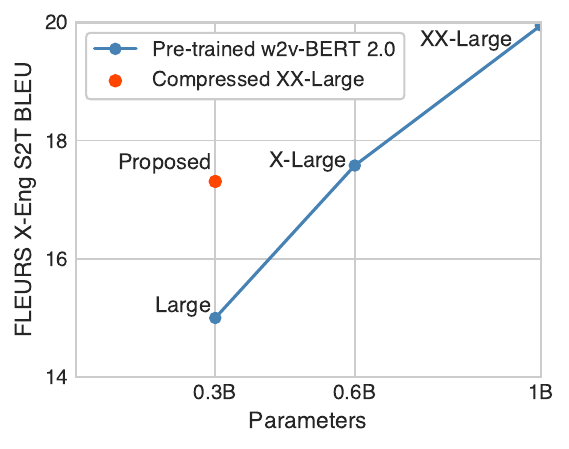}
    \vspace{-14pt}
    \caption{
        Encoder sizes vs. X-Eng speech-to-text translation BLEU scores.
        The proposed model is a compressed XX-Large model.
    }
    \label{fig:size-bleu}
    \vspace{-8pt}
\end{figure}

Although existing methods succeed in many tasks, most works focus on compressing small SSL models and evaluating with unrealistic problem setups.
Those works compress a HuBERT Base~\cite{hsu2021hubert} model~(95M parameters) to models around 20M to 30M parameters and evaluate with the Speech processing Universal PERformance Benchmark~(SUPERB)~\cite{yang2021superb,tsai-etal-2022-superb}.
These compressed models are unsuitable for complex tasks that require fine-tuning because of the small model capacities, limiting application scenarios.
Under this setting, the effectiveness of these methods for large-scale models and problems remains to be discovered.

To bridge the gap between academic research and real-world problems, we extend the speech encoder compression task to a large-scale pre-trained speech encoder (w2v-BERT 2.0~\cite{seamlessm4t2023}) and apply the compressed model to multilingual speech-to-text translation~(S2T).
This problem is challenging because the original model is significantly larger~(1B parameters), and the compressed model is fine-tuned with a more complicated yet realistic task.
Following previous studies, we use unlabeled data to compress an SSL pre-trained teacher model because this setup allows flexible utilization and avoids fine-tuning huge encoders.
Moreover, the compressed encoder has 300M parameters, which is currently the largest encoder size widely used in both production and academia~\cite{yang2021superb}.

Under this new problem setting, we propose Contrastive Layer-to-layer Distillation~(\proposed) by combining L2L KD~\cite{ashihara2022deep} and a contrastive masked prediction learning objective~\cite{baevski2020wav2vec2}.
First, some student model input frames are masked while the teacher remains unmasked.
Then, each masked student's hidden layer frame classifies the corresponding teacher's hidden layer frame from a set of distractors, where the distractors are randomly sampled from other frames of the teacher's representations.
After distillation, we evaluate the student model with internal and public benchmarks, covering S2T and multilingual ASR.
As shown in Fig.~\ref{fig:size-bleu} and Sec.~\ref{sec:exp}, \proposed~surpasses prior distillation methods, narrows the performance gap between large models~(0.6B and 1.0B parameters) and outperforms strong baselines like XLS-R~\cite{babu2022xlsr} and MMS~\cite{pratap2023mms}.

\begin{figure*}[t]
    \centering
    \includegraphics[width=0.82\linewidth]{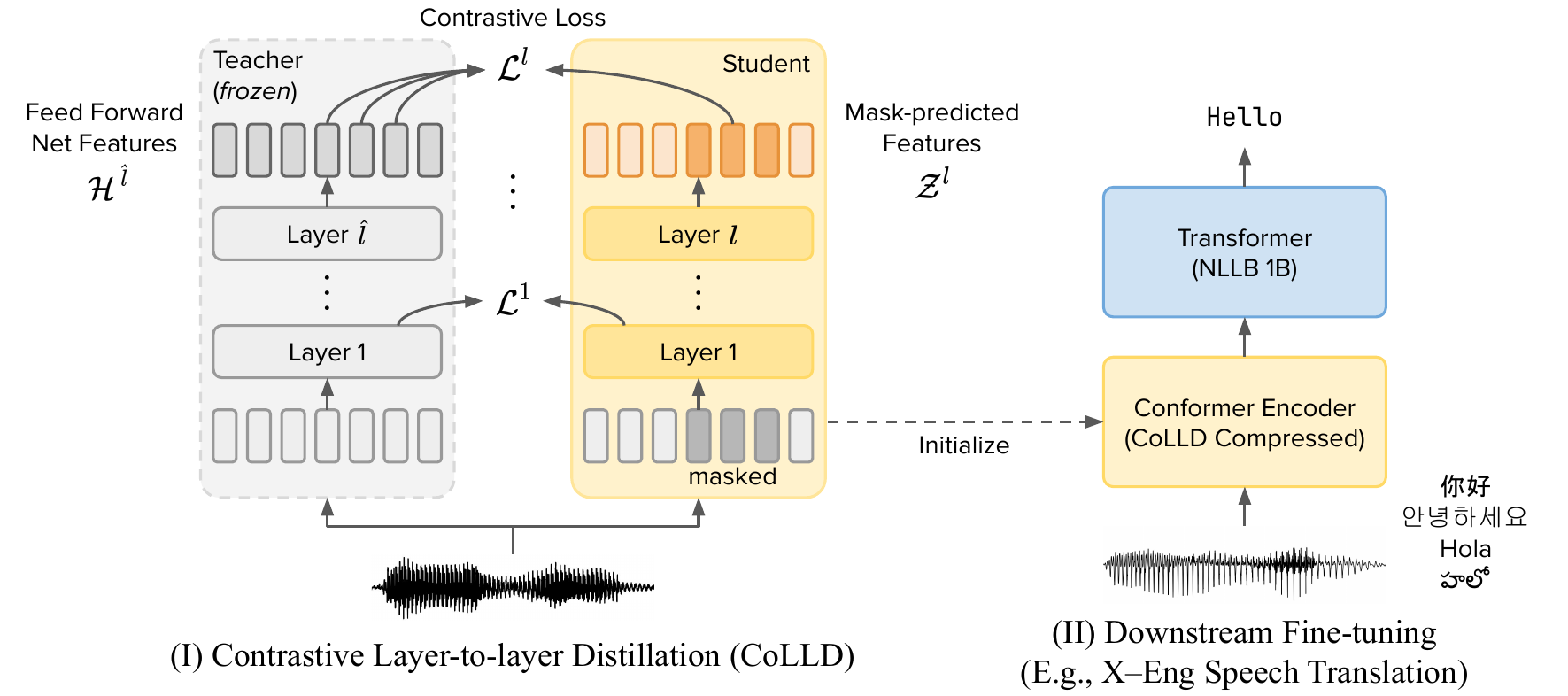}
    \vspace{-10pt}
    \caption{
        An illustration of the proposed Contrastive Layer-to-layer Distillation~(\proposed) framework.
        (I) \proposed~feeds the same input to a frozen teacher and a learnable student model, where the student's input frames are partially masked.
        For each student layer $l$, the masked representations learn to classify the corresponding teacher frame in layer $\hat{l}$ from $K$ distractor frames.
        (II) After distillation, the student model weights initialize downstream models and are fine-tuned with labeled data to perform tasks like multilingual speech translation.
    }
    \label{fig:framework}
    \vspace{-6pt}
\end{figure*}

\section{Method}
\label{sec:method}

\subsection{Overview}
\label{subsec:method-overview}

We propose the Contrastive Layer-to-layer Distillation~(\proposed) framework as shown in Fig.~\ref{fig:framework}.
First, the student's layers are trained to predict teacher hidden layer representations~(Sec.~\ref{subsec:method-l2l}).
Next, we incorporate masked prediction to encourage the student model to learn better representations~(Sec.~\ref{subsec:method-mlm}).
Finally, a contrastive learning objective prevents the model from collapsing.~(Sec.~\ref{subsec:method-contrast}).

\subsection{Layer-to-layer Distillation}
\label{subsec:method-l2l}

Moreover, as Ashihara et al.~\cite{ashihara2022deep} pointed out, deep and narrow student models better capture the teacher's behavior.
We follow \cite{lee2022fithubert} and \cite{ashihara2022deep} by assigning each student layer to predict a teacher's hidden layer.
The student-to-teacher layer mapping is obtained as follows.
Let $L^T$ and $L^S$ as the numbers of teacher and student layers, with $L^T \geq L^S$.
The $l^{\text{th}}$ student layer learns to predict the $\hat{l}^{\text{th}}$ teacher layer, where
\vspace{-5pt}
\begin{equation}
    \vspace{-5pt}
    \hat{l} = \text{round}\left( (l - 1) \frac{L^T - 1}{L^S - 1} \right) + 1,
    \label{eq:l2l}
\end{equation}
for $l = 1, 2, \dots, L^S$.
Each student layer is assigned to predict a unique teacher layer, and the selected layers are uniformly distributed across the teacher model.
This mapping rule allows flexible student architectures for different applications.

Previous works distill the final output of each teacher layer~\cite{chang2022distilhubert,lee2022fithubert}.
Inspired by data2vec~\cite{baevski2022data2vec}, we let the student model predict each teacher layer's feed-forward net~(FFN) features for better learning targets.
Specifically, the student learns from the outputs of the second FFN of each Conformer block in the teacher~\cite{gulati2020conformer}.

\subsection{Masked Prediction}
\label{subsec:method-mlm}

Prior KD methods usually keep the student's inputs unmasked~\cite{chang2022distilhubert,lee2022fithubert}, but many SSL methods rely on masked language modeling~\cite{baevski2020wav2vec2,hsu2021hubert,baevski2022data2vec}, and studies have shown this technique useful for knowledge distillation~\cite{wang2022lighthubert,jang2023recycle}.
Therefore, we only mask the student's input frames and apply L2L distillation to the masked frames.

\subsection{Contrastive Distillation Objective}
\label{subsec:method-contrast}
We found that utilizing L1 or L2 losses for KD sometimes leads to collapsed representations when incorporating masked prediction if the hyperparameters are not carefully tuned.
Hence, we propose a contrastive learning objective to mitigate this issue~\cite{oord2018cpc,baevski2020wav2vec2}.
For each masked timestep $t \in \mathcal{T}$ in an utterance, the student's $l^{\text{th}}$ layer output $\boldsymbol{z}^l_t$ predicts the $\hat{l}^{\text{th}}$ teacher layer representation $\boldsymbol{h}^{\hat{l}}_t$.
The student minimizes the distance between $\boldsymbol{z}^l_t$ and $\boldsymbol{h}^{\hat{l}}_t$.
The conventional L2 regression loss is written as
\vspace{-7pt}
\begin{equation}
    \vspace{-5pt}
    \mathcal{L}^l = \sum_{t \in \mathcal{T}} \left\| \boldsymbol{z}^l_t - \boldsymbol{h}^{\hat{l}}_t \right\|_2^2,
    \label{eq:l2}
\end{equation}
while the proposed contrastive distillation objective is
\vspace{-4pt}
\begin{equation}
    \vspace{-3pt}
    \mathcal{L}^{l} = -\sum_{t \in \mathcal{T}} \log \frac{\exp\left( \cos\left( \boldsymbol{z}^{l}_t, \boldsymbol{h}^{\hat{l}}_t \right) / \tau \right)}{\sum_{\boldsymbol{h}' \in \mathcal{H}^{\hat{l}}_t} \exp\left( \cos\left( \boldsymbol{z}^l_t, \boldsymbol{h}' \right) / \tau \right)},
    \label{eq:contrast}
\end{equation}
where $\mathcal{H}^{\hat{l}}_t$ is a set composed of $\boldsymbol{h}^{\hat{l}}_t$ and $K$ distractors~\cite{oord2018cpc} sampled from the $\hat{l}^{\text{th}}$ teacher layer with indices also in $\mathcal{T}$.
$\tau > 0$ is a hyperparameter and $\cos(\cdot,\cdot)$ denotes cosine similarity.
With this objective, the model is expected to avoid collapsing.

\begin{table*}[t]
    \centering
    \caption{
        BLEU scores of multilingual speech-to-text translation~(X-Eng S2T) evaluated on CoVoST 2~\cite{wang2021covost2} and FLEURS 101 languages test set~\cite{conneau2023fleurs}.
        Excluding pre-trained from scratch toplines, each model has 0.3B parameters.
        Avg indicates an averaged score across all languages.
    }
    \label{tab:ft-s2t-asr}
    \vspace{1pt}
    \begin{tabular}{@{~~}l@{~~~~}c@{~~~}c@{~~~}c@{~~~}c@{~~}c@{~~}c@{~~~}c@{~~~}c@{~~~}c@{~~~}c@{~~~}c@{~~~}c@{~~~}c@{~~~}c@{~~~}c@{~~}}
        \toprule
        & \multicolumn{4}{@{~~~}c@{~~}}{CoVoST 2} & & \multicolumn{8}{@{~~}c@{~~~}}{FLEURS-101} \\
        \cmidrule{2-5}
        \cmidrule{7-14}
        Method & \footnotesize{High} & \footnotesize{Mid} & \footnotesize{Low} & \footnotesize{Avg} & & \footnotesize{WE} & \footnotesize{EE} & \footnotesize{CMN} & \footnotesize{SSA} & \footnotesize{SA} & \footnotesize{SEA} & \footnotesize{CJK} & \footnotesize{Avg} \\
        \midrule
        
        \multicolumn{3}{@{~~}l@{~~~}}{\textbf{Pre-trained w2v-BERT 2.0}} \\
        ~~~XX-Large (1.0B Teacher) & 37.6 & 35.8 & 29.7 & 32.6 & & 27.5 & 25.8 & 20.2 & 10.0 & 19.1 & 16.2 & 14.3 & 20.0 \\
        ~~~X-Large (0.6B) & 36.4 & 34.0 & 28.0 & 31.0 & & 25.7 & 23.7 & 17.0 & 8.0 & 16.4 & 13.9 & 9.9 & 17.6 \\
        ~~~Large\textsubscript{12} (0.3B) & 33.5 & 31.3 & 23.7 & 27.4 & & 22.4 & 20.7 & 14.3 & 7.1 & 12.9 & 11.4 & 8.1 & 15.0 \\

        \midrule
        \multicolumn{3}{@{~~}l@{~~~}}{\textbf{Layer Removal from Pre-trained XX-Large}} \\
        ~~~Layer Skipping Large\textsubscript{12} & 29.6 & 26.4 & 15.9 & 21.0 & & 15.6 & 14.0 & 9.1 & 3.5 & 8.2 & 7.1 & 4.2 & 9.7 \\
        ~~~Bottom Layers Large\textsubscript{12} & 31.4 & 28.9 & 21.8 & 25.3 & & 20.0 & 18.1 & 12.4 & 5.9 & 10.8 & 10.2 & 6.9 & 13.1 \\ 
        
        \midrule
        \multicolumn{3}{@{~~}l@{~~~}}{\textbf{\proposed~from Pre-trained XX-Large}} \\
        ~~~Large\textsubscript{12} & 34.6 & 32.8 & 25.5 & 29.0 & & \textbf{24.3} & 23.2 & 17.3 & 7.5 & 16.2 & \textbf{14.0} & \textbf{12.3} & 17.2 \\
        ~~~Large\textsubscript{40} & \textbf{35.4} & \textbf{33.6} & \textbf{26.9} & \textbf{30.1} & & 24.0 & \textbf{23.3} & \textbf{17.9} & \textbf{7.6} & \textbf{16.6} & \textbf{14.0} & 11.8 & \textbf{17.3} \\
        ~~~~~~Loss \hfill Contrastive~(Eq.~\ref{eq:contrast}) $\rightarrow$ L2~(Eq.~\ref{eq:l2}) & 33.7 & 32.2 & 25.2 & 28.5 & & 22.6 & 22.0 & 16.8 & 6.9 & 15.1 & 13.3 & 11.2 & 16.2 \\
        ~~~~~~Target \hfill FFN $\rightarrow$ Layer Output & 34.1 & 32.2 & 25.5 & 28.7 & & 22.7 & 22.0 & 16.8 & 7.4 & 15.9 & 13.3 & 10.9 & 16.4 \\
        ~~~~~~Data \hfill Multilingual $\rightarrow$ Monolingual & 33.7 & 31.7 & 24.8 & 28.2 & & 22.4 & 21.4 & 15.1 & 5.8 & 13.8 & 11.8 & 9.5 & 15.2 \\
        ~~~~~~Initialization~~~~ \hfill Random $\rightarrow$ Layer Skipping Large\textsubscript{12} & 30.1 & 27.3 & 18.0 & 22.5 & & 19.1 & 17.6 & 11.7 & 4.8 & 10.7 & 9.1 & 6.9 & 12.4 \\
        ~~~~~~Initialization \hfill Random $\rightarrow$ Bottom Layers Large\textsubscript{12} & 29.1 & 25.8 & 17.8 & 21.8 & & 18.6 & 17.0 & 11.3 & 4.9 & 10.3 & 8.7 & 6.3 & 12.0 \\

        \midrule
        \multicolumn{3}{@{~~}l@{~~~}}{\textbf{\proposed~from Fine-tuned XX-Large}} \\
        ~~~Large\textsubscript{40} + S2T Fine-tuned XX-Large Teacher & 36.2 & 34.5 & 27.9 & 31.0 & & 25.2 & 24.6 & 19.1 & 8.2 & 18.2 & 15.4 & 13.7 & 18.5 \\
        
        \bottomrule
    \end{tabular}
    \vspace{-5pt}
\end{table*}

\begin{table}[t]
    \centering
    \caption{
        w2v-BERT 2.0~\cite{seamlessm4t2023} architectures with different dimensions, feed-forward net sizes~(FFN), and attention heads.
        The number of parameters~(Param) and multiply–accumulate operation~(MACs) during forward pass indicate required spaces and computation costs.
        MACs are calculated with an input utterance of 20 seconds long.
}
    \label{tab:arch}
    \vspace{1pt}
    \begin{threeparttable}
    \begin{tabular}{@{~~}l@{~~}c@{~~~}c@{~~~}c@{~~~}c@{~~~}c@{~~~}c@{~~}}
        \toprule
        Model & Param & Dim & FFN & Layer & Head & GMACs$\downarrow$ \\
        \midrule
        XX-Large & 1.0B & 1024 & 4096 & 40 & 16 & 1214.1 \\
        X-Large & 0.6B & 1024 & 4096 & 24 & 16 & 728.5 \\
        Large\textsubscript{12} & 0.3B & 1024 & 4096 & 12 & 16 & 364.3 \\
        Large\textsubscript{40} & 0.3B & 768 & 1024 & 40 & 8 & 462.3 \\
        \bottomrule
    \end{tabular}
    \begin{tablenotes}[flushleft]
        \item 
            {\footnotesize MACs computation: {https://github.com/zhijian-liu/torchprofile}}
    \end{tablenotes}
    \end{threeparttable}
    \vspace{-10pt}
\end{table}

\section{Experiments}
\label{sec:exp}

\subsection{Setup}
\label{subsec:exp-setup}

\subsubsection{Model}
\label{subsubsec:exp-model}

All experiments are based on w2v-BERT 2.0~\cite{seamlessm4t2023}, a series of SSL speech encoders trained with contrastive learning~\cite{baevski2020wav2vec2} and masked language modeling~\cite{hsu2021hubert}.
The Conformer~\cite{gulati2020conformer} architectures and forward computing costs are listed in Table~\ref{tab:arch}.
A depth-wise convolution kernel size of 31 is used.
Each model takes 80-dimensional filter bank features as input and downsamples each utterance by concatenating consecutive frames to reduce the frame rate from 100Hz to 50Hz.
Excluding Large\textsubscript{40}, all w2v-BERT 2.0 models are pre-trained from scratch with an internal corpus containing 4M hours of unlabeled speech, covering 143+ languages.
Unless stated otherwise, students are randomly initialized Large\textsubscript{40} or Large\textsubscript{12} models that distill knowledge from the XX-Large teacher.

\subsubsection{Knowledge Distillation}
\label{subsubsec:exp-kd-impl}

\begin{table*}[t]
    \caption{
        SSL pre-trained models with 0.3B parameters on the 10-minute set of the ML-SUPERB benchmark~\cite{shi2023mlsuperb}.
        The metrics include accuracy~(Acc\%), character error rate~(CER\%), phone error rate~(PER\%), and SUPERB score~(SUPERB$_s$)~\cite{feng2022superb}.
    }
    \vspace{1pt}
    \label{tab:ml-superb}
    \centering
    \begin{tabular}{@{~~}l@{~~~}c@{~~~}c@{~~~}c@{~~~}c@{~~~}c@{~~~}c@{~~~}c@{~~~}c@{~~~}c@{~~~}c@{~~}}
        \toprule
        & \multicolumn{2}{@{~~~}c@{~~~}}{\multirow{2}{*}{\shortstack{~\\Pre-training / \\Distillation Data}}} & \multirow{2}{*}{Mono-ASR} & \multicolumn{2}{@{~~~}c@{~~~}}{Multi-ASR} & LID & \multicolumn{3}{@{~~~}c@{~~~}}{Multi-ASR + LID} &   \\
        \cmidrule{5-6}
        \cmidrule{8-10}
        & & & & Normal & Few-shot & Normal & \multicolumn{2}{@{~~~}c@{~~~}}{Normal} & Few-shot \\
        \cmidrule{2-3}
        \cmidrule{8-9}
        SSL Model & \#Hours & \#Langs & CER/PER$\downarrow$ & CER$\downarrow$ & CER$\downarrow$ & Acc$\uparrow$ & Acc$\uparrow$ & CER$\downarrow$ & CER$\downarrow$ & SUPERB$_s$$\uparrow$ \\
        \midrule
        \multicolumn{3}{@{~~}l@{~~~}}{\textbf{No Compression Baseline}} \\
        ~~~~XLSR 53~\cite{conneau2020xlsr} & 56k & 53 & 49.5 & 33.9 & 43.6 & 6.6 & 45.6 & 33.4 & 43.2 & 403.4 \\
        ~~~~XLS-R 128~\cite{babu2022xlsr} & 400k & 128 & 39.7 & 29.2 & 40.9 & 66.9 & 55.6 & 28.4 & 42.1 & 734.1 \\
        ~~~~MMS~\cite{pratap2023mms} & 491k & 1406 & \textbf{33.8} & 28.7 & 36.5 & 62.3 & 71.9 & 31.5 & 30.9 & 829.1 \\
        ~~~~w2v-BERT 2.0 Large\textsubscript{12} & 4M & 143+ & 46.6 & 27.2 & 32.2 & 37.0 & 78.5 & 27.2 & 31.7 & 698.8 \\
        \midrule
        \multicolumn{3}{@{~~}l@{~~~}}{\textbf{Proposed}} \\
        ~~~~CoLLD Large\textsubscript{40} & 92k & 143+ & 35.5 & \textbf{22.2} & \textbf{29.6} & \textbf{82.8} & \textbf{85.7} & \textbf{21.9} & \textbf{28.7} & \textbf{988.7} \\
        \bottomrule
    \end{tabular}
    \vspace{-6pt}
\end{table*}

We implement experiments with fairseq~\cite{ott2019fairseq}.
Only 92k hours of audio data in the 4M hours corpus are used for distillation because KD requires fewer updates than pre-training, where the amount of used training data is calculated according to~\cite{chang2023spin}.
We set $\tau=$ 0.1 and $K=$ 100 in Eq.~\ref{eq:contrast}.
Downsampled features are randomly masked with a span of 10 frames and a probability of 0.065, resulting in approximately 49\% of masked frames.
Each model is trained with 200k updates using an Adam optimizer~\cite{kingma2014adam} with a peak learning rate of 10\textsuperscript{$-$4}, $\beta_1 =$ 0.9, $\beta_2 =$ 0.98, $\epsilon=$ 10\textsuperscript{$-$6}, and a weight decay of 10\textsuperscript{$-$2}.
The learning rate ramps up linearly in the first 4k updates and linearly decays to 0 for the rest.
Each model is compressed on 32 NVIDIA A100 80GB GPUs, with an effective batch size of 27.7 minutes of audio data in each update.
Large\textsubscript{12} and Large\textsubscript{40} students take 2 and 4 days to distill from the XX-Large teacher.
Although the parameters of 0.3B models are similar, the distillation time of the 40-layer student is higher because the forward operation of each hidden layer cannot be parallelized.
Some prior KD methods are not included for comparison because they require complex implementation and hyperparameter search.

\subsubsection{Multilingual Speech Translation}
\label{subsubsec:exp-s2t-impl}

The speech-to-English-text translation~(X-Eng S2T) model comprises a Conformer encoder, a length adaptor~\cite{zhao2022madapter}, and a 1.3B-parameter NLLB-200 machine translation model~\cite{costa2022nllb}.
The fine-tuning data include approximately 60k hours of paired speech and translation text that cover 88 X-English directions.
The Conformer encoder is fine-tuned entirely, but only the layer norm and self-attention for NLLB.
The learning rate linearly increases to 10$^{-\text{4}}$ in the first 5k updates (2 $\times$ 10$^{-\text{4}}$ for XX-Large), and then follows the inverse square root schedule~\cite{vaswani2017attention}.
All models are trained with an effective batch size of 64 minutes of audio and 150k updates.
We use 16 to 64 NVIDIA V100 32GB GPUs, depending on the model size.
We evaluate fine-tuned S2T models on CoVoST 2~\cite{wang2021covost2} and FLEURS~\cite{conneau2023fleurs} with a decoding beam size of 5.

\subsection{Fine-tuning Results}
\label{subsec:exp-ft}

This section reveals the effectiveness of \proposed~through fine-tuning w2v-BERT 2.0 models on X-Eng S2T.
As shown in Table~\ref{tab:ft-s2t-asr}, we offer three pre-trained from scratch w2v-BERT 2.0 models, where the 0.3B model is served as a baseline.  
Layer removal baselines preserve 30\% of the layers of the XX-Large model by either preserving the bottom layers or uniformly skipping layers following Eq.~\ref{eq:l2l}.

\proposed~Large\textsubscript{40} surpasses 0.3B baselines by at least two BLEUs in most subsets, indicating that the student successfully acquires knowledge from the XX-Large teacher.
Although \proposed~Large\textsubscript{40} is incapable of reaching the same performance as the 1.0B teacher because of the model capacity, the gap between the 0.3B and 0.6B models is significantly reduced.
Especially in FLEURS, \proposed~offers slightly superior BLEU scores in most subsets compared with the 0.6B topline.
Hence, \proposed~Large\textsubscript{40} is comparable with the X-Large w2v-BERT 2.0 but requires only half of the parameters.

We offer ablation studies in the same table.
The overall S2T performance is degraded by replacing each of the proposed components in \proposed~with prior methods, indicating the necessity of the design of \proposed.
First, a shallow and wide student architecture~(Large\textsubscript{12}) drops one BLEU score in most test sets compared with the deeper model~(Large\textsubscript{40}), corroborating with prior studies~\cite{ashihara2022deep,lee2022fithubert}.
Still, Large\textsubscript{12} outperforms all baselines, and the fine-tuning and inference costs of the shallow model are lower than those of the deep model.
Therefore, the choice between shallow and deep models depends on the application scenario.
Second, optimizing with L2 loss or learning from each teacher layer's output leads to 1 to 2 BLEU score degradation, showing that the proposed techniques distill better representations from the teacher.
Third, replacing distillation data with a 1k hours English speech corpus decreases BLEU scores but performs better than the baselines, implying that \proposed~still works even when the training data diversity is reduced.
Furthermore, initializing student models with some teacher layers results in significantly worse scores, so model initialization is unnecessary.
Note that we do not compare with DistilHuBERT because prior works have shown L2L KD has superior performance~\cite{ashihara2022deep,lee2022fithubert}.
The ablation studies clearly show the importance of the proposed \proposed.

To push the limit of \proposed, we consider distilling from an S2T fine-tuned teacher for comparison.
In the last part of Table~\ref{tab:ft-s2t-asr}, the results of a \proposed~Large\textsubscript{40} model distilled from an S2T fine-tuned XX-Large teacher are reported.
This compressed model offers superior performance compared with the 0.6B topline in many evaluation subsets, showing that \proposed~is applicable to both pre-trained and fine-tuned w2v-BERT 2.0 models.
Thus, if a teacher model fine-tuned with labeled data is available, \proposed~produces better-compressed models.
Overall, \proposed~successfully compresses a pre-trained XX-Large w2v-BERT 2.0 by 70\% while retaining good X-Eng S2T performance.

\subsection{Multilingual SUPERB}
\label{subsec:exp-ml-superb}

This section evaluates \proposed~with Multilingual SUPERB~(ML-SUPERB)~\cite{shi2023mlsuperb}, a standard multilingual speech processing benchmark, to offer a more comprehensive comparison with other SSL models.
ML-SUPERB covers 143 languages and four tasks: monolingual ASR~(Mono-ASR), multilingual ASR~(Multi-ASR), language identification~(LID), and Multi-ASR + LID.
We use the 10-minute set of ML-SUPERB to show the performance of pre-trained models in a low-resource setting.
For a fair comparison, the pre-trained and distilled models are frozen and serve as feature extractors during downstream model training.
We follow the implementation as in ESPnet~\cite{watanabe2018espnet}.

As shown in Table~\ref{tab:ml-superb}, w2v-BERT 2.0 offers a solid baseline compared to prior works because this model is trained with significantly more data.
Next, \proposed~surpasses w2v-BERT 2.0 and other prior methods in most ML-SUPERB tasks and achieves the best overall SUPERB score by using only 92k hours of distillation data.
The results again corroborate that \proposed~successfully distills knowledge from the XX-Large teacher.

\subsection{Impact of Distillation Updates}
\label{subsec:exp-updates}

\begin{figure}[t]
    \centering
    \includegraphics[width=0.75\linewidth]{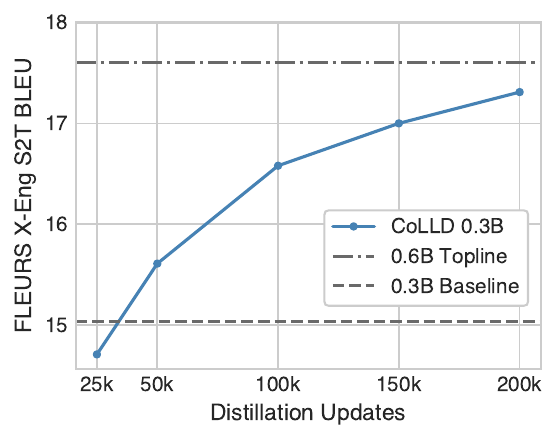}
    \vspace{-12pt}
    \caption{
        Distillation updates vs. FLEURS-101 X-Eng BLEU scores.
    }
    \label{fig:update-bleu}
    \vspace{-5pt}
\end{figure}

This section investigates the impact of the data required for \proposed~by varying the total number of distillation updates.
As shown in Fig.~\ref{fig:update-bleu}, \proposed~surpasses the 0.3B pre-trained from scratch baseline with only 50k of distillation updates.
Meanwhile, when trained with 200k updates, \proposed~reaches a similar performance as the 0.6B topline model.
Therefore, the amount of distillation data is highly correlated to downstream performance, and the distilled models offer better representations when more data and computation resources are available.

\section{Conclusion}
\label{sec:conclusion}

This paper proposes \proposed, a novel model compression method by combining layer-to-layer knowledge distillation and contrastive learning for large-scale multilingual speech encoders.
We show that \proposed~is superior over prior compression methods on multilingual speech recognition and speech-to-text translation by evaluating the proposed methods on internal and public benchmarks.
This approach reduces model sizes of powerful pre-trained speech encoders while retaining good performance after fine-tuning, enabling on-device and streaming applications.

\clearpage

\bibliographystyle{IEEEtran}
\bibliography{refs}

\begin{figure*}[t]
    \centering
    \includegraphics[width=\linewidth]{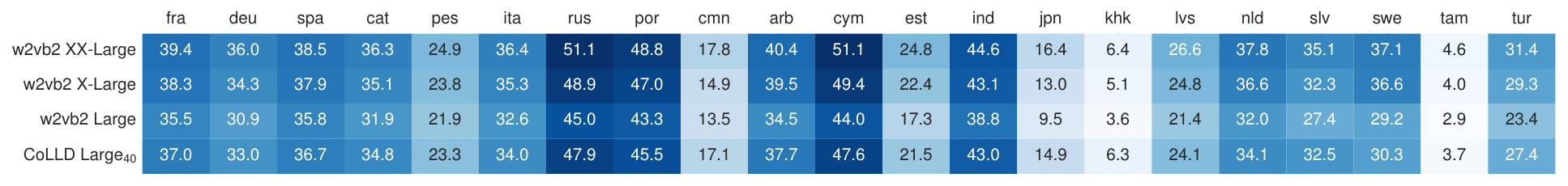}
    \vspace{-24pt}
    \caption{Complete BLEU scores on the CoVoST 2 X-Eng S2T task. w2vb2 denotes w2v-BERT 2.0.}
    \label{fig:res-s2t-cvst2}
    \vspace{-6pt}
\end{figure*}
\begin{figure*}[t]
    \centering
    \includegraphics[width=\linewidth]{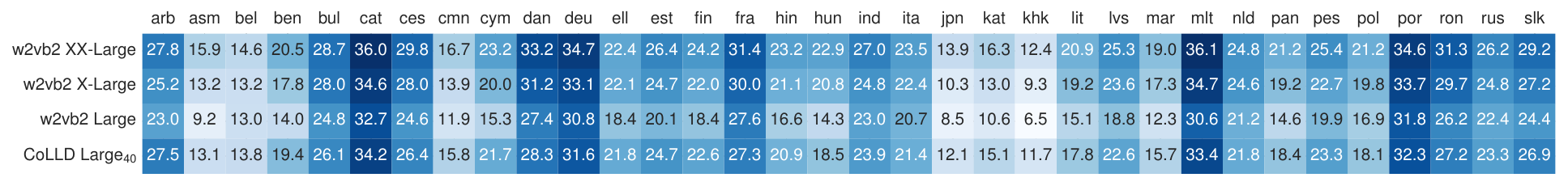} \\[-6pt]
    \includegraphics[width=\linewidth]{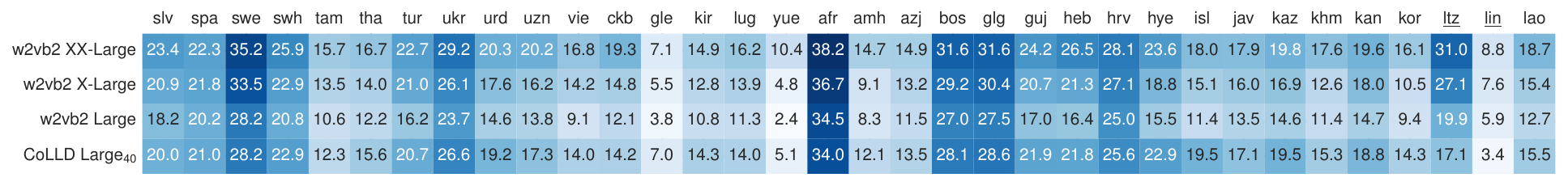} \\[-6pt]
    \includegraphics[width=\linewidth]{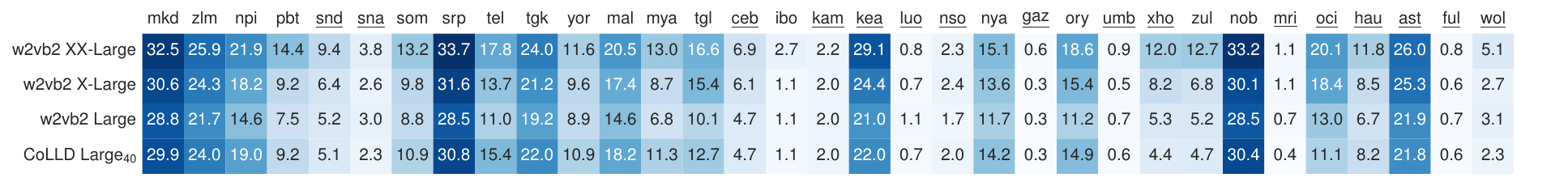}
    \vspace{-20pt}
    \caption{Complete BLEU scores on the FLEURS-101 X-Eng S2T task. w2vb2 denotes w2v-BERT 2.0. Underlined languages indicate unseen languages in X-Eng fine-tuning data.}
    \label{fig:res-s2t-fleurs}
    \vspace{-15pt}
\end{figure*}

\section{Appendix}
\label{sec:appendix}

\subsection{Knowledge Distillation Details}
\label{subsec:app-kd-detail}

\begin{table}[t]
    \centering
    \caption{
        The $l^{\text{th}}$ student layer to the $\hat{l}^{\text{th}}$ teacher layer mapping for CoLLD derived from Eq.~\ref{eq:l2l} when distilling from a 1B teacher.
    }
    \label{tab:l2l-map}
    \vspace{2pt}
    \begin{tabular}{@{~~}l@{~~~}c@{~~~}c@{~~~}l@{~~}}
        \toprule
        Architecture & $L^S$ & $L^T$ & $( l, \hat{l})$ \\
        \midrule
        \multirow{3}{*}{Large\textsubscript{12}} & \multirow{3}{*}{12} & \multirow{3}{*}{40} & (1, 1), (2, 5), (3, 8), (4, 12), \\
        & & & (5, 15), (6, 19), (7, 22), (8, 26), \\
        & & & (9, 29), (10, 33), (11, 36), (12, 40) \\
        \midrule
        Large\textsubscript{40} & 40 & 40 & (1, 1), (2, 2), (3, 3), $\dots$, (40, 40) \\
        \bottomrule
    \end{tabular}
    \vspace{-4pt}
\end{table}

Here, we offer details about the knowledge distillation implementation.
In Table~\ref{tab:l2l-map}, we show the student-to-teacher layer mapping in our distillation experiments.
Next, the L2 regression loss for an utterance can be expressed as 
\vspace{-8pt}
\begin{equation}
    \vspace{-5pt}
    \mathcal{L}_{\ell_2} = \frac{1}{DL^S|\mathcal{T}|}\sum_{l=1}^{L^S} \sum_{t \in \mathcal{T}} \left\| \boldsymbol{z}^l_t - \boldsymbol{h}^{\hat{l}}_t \right\|_2^2,
\end{equation}
where $D$ is the dimension of the representations $\boldsymbol{z}$ and $\boldsymbol{h}$, and $|\mathcal{T}|$ is the number of masked time steps.
For contrastive learning, the loss function is
\vspace{-8pt}
\begin{equation}
    \mathcal{L}_{\text{Contrastive}} = -\frac{1}{L^S|\mathcal{T}|} \sum_{l=1}^{L^S} \sum_{t \in \mathcal{T}} \log \frac{\exp\left( \cos\left( \boldsymbol{z}^{l}_t, \boldsymbol{h}^{\hat{l}}_t \right) / \tau \right)}{\sum_{\boldsymbol{h}' \in \mathcal{H}^{\hat{l}}_t} \exp\left( \cos\left( \boldsymbol{z}^l_t, \boldsymbol{h}' \right) / \tau \right)}.
\end{equation}
Finally, the losses of all utterances within a mini-batch are averaged to obtain the total loss function for optimization.

\subsection{S2T Fine-tuning Details}
\label{subsec:app-ft-detail}

This section offers implementation details of X-Eng S2T fine-tuning.
Some fine-tuning hyperparameters for different model architectures are shown in Table~\ref{tab:hp-s2t}.
First, the maximum length of an input utterance is 30 seconds, and the maximum number of output tokens is 113.
Second, the input frames are randomly masked similar to the distillation process, but with a mask length of 5 and a masking probability of 0.02.
Next, layer dropping of probability 0.1 is applied to both w2v-BERT 2.0 and NLLB models.
Moreover, the NLLB transformer model is pre-trained with machine translation tasks, which take text as input, so we add a length adaptor~\cite{zhao2022madapter} after the speech encoder to match the sequence length between speech and text.
The adaptor begins with a 1-D CNN layer (kernel size $=$ stride $=$ 8) and a gated linear unit, followed by a single Conformer encoder layer with a convolution kernel size of 31.
After this adaptor, the utterance length is reduced by a factor of eight to match the text modality.

\begin{table}[t]
    \centering
    \caption{X-Eng S2T fine-tuning hyperparameters for different model architectures.}
    \label{tab:hp-s2t}
    \vspace{2pt}
    \begin{tabular}{l@{~~~}c@{~~~}c@{~~~}c@{~~~}c}
        \toprule
        & \multirow{2}{*}{\shortstack{Learning\\Rate}} & \multirow{2}{*}{\shortstack{Batch Size\\Per GPU}} & \multirow{2}{*}{\shortstack{Gradient\\Accumulation}} & \multirow{2}{*}{\shortstack{GPUs}} \\
        Model &  &  &  \\
        \midrule
        XX-Large & 2 $\times$ 10$^{-\text{4}}$ & 30 sec & 2 & 64 \\
        X-Large & 1 $\times$ 10$^{-\text{4}}$ & 60 sec & 2 & 32 \\
        Large\textsubscript{12} & 1 $\times$ 10$^{-\text{4}}$ & 60 sec & 2 & 32 \\
        Large\textsubscript{40} & 1 $\times$ 10$^{-\text{4}}$ & 48 sec & 2 & 40 \\
        \bottomrule
    \end{tabular}
    \vspace{-4pt}
\end{table}

\subsection{Complete X-Eng S2T Results}
\label{subsec:app-s2t-res}

In Fig.~\ref{fig:res-s2t-cvst2} and \ref{fig:res-s2t-fleurs}, we show the BLEU scores of several models of all languages in the CoVoST 2 and FLEURS evaluation sets.
The details of different languages in the fine-tuning dataset can be found in Table 35 of~\cite{seamlessm4t2023}.
Most unseen languages in the FLEURS testing sets have low BLEU scores.
However, some unseen languages like \texttt{ast}~(Asturian) and \texttt{ltz}~(Luxembourgish) have high BLEU scores.
We suspect high-resource languages in the same language family cause this phenomenon.

\end{document}